# Surrogate-assisted level-based learning evolutionary search for heat extraction optimization of enhanced geothermal system

Guodong Chen[1]; Xin Luo[1]; Chuanyin Jiang[2]; Jiu Jimmy Jiao[1,*]

1 Department of Earth Sciences, The University of Hong Kong, Hong Kong, China

2 HSM, University of Montpellier, CNRS, IRD, Montpellier, France

Corresponding Author(s): Jiu Jimmy Jiao (jjiao@hku.hk).

## Abstract

An enhanced geothermal system is essential to provide sustainable and long-term geothermal energy supplies and reduce carbon emissions. Optimal well-control scheme for effective heat extraction and improved heat sweep efficiency plays a significant role in geothermal development. However, the optimization performance of most existing optimization algorithms deteriorates as dimension increases. To solve this issue, a novel surrogate-assisted level-based learning evolutionary search algorithm (SLLES) is proposed for heat extraction optimization of enhanced geothermal system. SLLES consists of classifier-assisted level-based learning pre-screen part and local evolutionary search part. Specifically, the classifier-assisted level-based learning strategy employs probabilistic neural network as classifier to classify the offspring into pre-set number of levels. The offspring in different levels uses level-based learning strategy to generate more promising and informative candidates pre-screened by classifier to conduct real simulation evaluations. In the local evolutionary search part, a surrogate model is constructed at the local promising area. The optima of the surrogate model obtained by the optimizer is selected to conduct real simulation evaluations. The cooperation of the two parts has realized the balance between the exploration and exploitation during the optimization process. After iteratively sampling from the design space, the robustness and effectiveness of the algorithm are proven to be improved significantly. To the best of our knowledge, the proposed algorithm holds state-of-the-art simulation-involved optimization framework. Comparative experiments have been conducted on benchmark functions, a two-dimensional fractured reservoir and a three-dimensional enhanced geothermal system. The proposed algorithm outperforms other five state-of-the-art surrogate-assisted algorithms on all selected benchmark functions. The results on the two heat extraction cases also demonstrate that SLLES can achieve superior optimization performance compared with traditional evolutionary algorithm and other surrogate-assisted algorithms. This work lays a solid basis for efficient geothermal extraction of enhanced geothermal system and sheds light on the model management strategies of data-driven optimization in the areas of energy exploitation.

# 1. Introduction

Geothermal energy from geothermal reservoirs is regarded as one of the best alternative energy sources to hydrocarbon resources due to its renewable and high potential properties [1-3]. Enhanced geothermal system (EGS) is able to provide sustainable and long-lasting geothermal energy supplies, reduce carbon emissions and mitigate global warming [4]. However, the development of EGS has encountered many serious challenges in technical and economical perspectives. As such, many geothermal reservoirs are abandoned for the low-production and risks derived from geological uncertainties and complexity [2]. Besides, many EGSs are commercially infeasible due to issues such as thermal breakthrough and water loss in geothermal development [5]. Thus, how to obtain optimal geothermal development to achieve efficient heat extraction is of great significance for geothermal exploitation.

As the prerequisite for predicting groundwater flow and thermal regimes in geothermal reservoirs, forward numerical simulation, including finite element method, finite volume method, and finite difference method, etc., has been widely used to describe the interaction between subsurface fluids and rocks and to understand geothermal distribution [6, 7]. Fractured-rock modeling, which has been commonly employed in EGS, is usually classified into three categories: equivalent porous medium model [8, 9], dual-porosity model [10-12], and discrete fracture model (DFM) [13-15]. Noorishad and Mehran [8] treated fractured aquifers as equivalent porous medium to simulate solute transport. Gerke and Van Genuchten [12] developed dual-porosity model to study variably saturated water flow and solute transport in fractured aquifers. Cacas et al. [15] presented a stochastic fracture network model with impermeable matrix to simulate flow regimes in the fractured networks. DFM builds the fractures explicitly and is able to simulate flow and solute transport more realistically. However, it suffers from time-consuming pre-processing of unstructured grid [16, 17]. In recent years, embedded discrete fracture model (EDFM) [18-21] has been proposed and gained considerable attention. EDFM treats fracture and matrix as two separate domains, and inserts fractures into matrix with structured grid. Moinfar et al. [21] developed an EDFM for 3D compositional reservoir building upon a finite difference reservoir simulator. Tripoppoom et al. [19] employed EDFM to characterize hydraulic and natural fracture parameters in shale reservoir with assisted history matching. EDFM is computationally efficient, but the trade-off is that it loses some accuracy near the fractures, since the grid size of the matrix that near the fractures in EDFM is much larger than in DFM.

In the past decades, researchers have made great efforts to improve the modeling accuracy and enhance the prediction of fluid flow dynamics when characterizing the EGS [22]. Commonly used investigation techniques for subsurface system inversions are hydraulic tomography [23], stress-based tomography [24], tracer tests [25], thermal experiments [13] and geophysical signals [26]. Wu et al. [25] inversed the tracer data to estimate the spatial distributions of apertures in the fractured model, using principal component analysis and ensemble smoother with multiple data assimilation. Vogt et al. [27] applied the tracer test to predict the long-term performance of the EGS using an Ensemble Kalman Filter based data assimilation method. Afshari Moein et al. [24] introduced stress-based tomography to characterize the fracture network and employed Markov Chain Monte Carlo sequence to update fracture network realizations. Zhou et al. [13] proposed an inversion method accelerated with deep neural network to infer the fracture density and fractal dimension of 2D fracture fields. Ringel et al. [14] introduced a Markov Chain Monte Carlo based 3D fracture network stochastic characterization to identify variable fracture locations and orientations. These more reliable geological models are the basis for the geothermal development design and optimization to maximize the heat recovery [28].

Various optimization algorithms, which can be mainly divided into gradient-based algorithm and derivative-free algorithm, have been developed to solve geothermal optimization problems [2, 29]. Gradient-based algorithms use adjoint-based method to calculate the gradient of objective function involving simulation. Gradient-based algorithms are computationally efficient, yet are easy to trap into a local optima [30]. Approximate gradient-based algorithms, such as simultaneous perturbation stochastic approximation [31] and ensemble optimization, use approximate gradient to update the solution, yet are still prone to trapping into a local optima. Heuristic algorithms, e.g., genetic algorithm (GA) [32], particle swarm optimization (PSO) [33] and differential evolution (DE) [34], have broad applications in many fields due to their high feasibility in jumping out from the local optima [32]. However, these algorithms require a large number of simulation evaluations to converge. Surrogate-assisted evolutionary algorithms (SAEAs) have gained increasing attention in subsurface geothermal systems in recent years due to the promising performance in reducing computational cost during optimization processes [35]. Surrogate model is a computationally efficient mathematical model to approximate the input/output relationship of complex systems [36, 37]. Various machine learning methods have been used as surrogate models, such as polynomial regression surface [38], radial basis function (RBF) [33, 39], support vector machine, Gaussian process (also known as Kriging model), and artificial neural network. For a clear understanding of the states of machine learning and optimization techniques in geothermal resource development and a concise comparison among these algorithms, a thorough review of recent studies has been conducted as shown in details in **Table 1**. Asai et al. [38] employed second order polynomial function as surrogate model to investigate the sensitivity of several factors on the produced water temperature and the heat recovery. Samin et al. [40] adopted multi-objective genetic algorithm to maximize long-term performance of geothermal reservoirs while minimizing cost. Pollack and Mukerji [5] used particle swarm optimization to maximize the net present value (NPV) of EGS by considering subsurface uncertainty. Chen et al. [36] used multivariate adaptive regression spline as surrogate model to achieve the optimal design of a potential geothermal reservoir. However, when handling high-dimensional optimization problems, surrogate-based optimization algorithms always suffer from "the curse of dimensionality". Moreover, these existing algorithms mainly depend on offline surrogate without further infilling samples. Therefore, there is an urgent need to promote effective surrogate model featured by online infilling and the alleviation of "the curse of dimensionality".

**Table 1.** Summary of machine learning and optimization techniques for geothermal resource development in recent years and comparison with those used in this study.

| | Decision variables | Optimization type | Surrogate | Surrogate type | Optimizer | Dimension | Additional information |
|---|---|---|---|---|---|---|---|
| Chen et al. [36] | Injection-production design | Single objective | Multivariate adaptive regression spline | Offline | Quadratic approximation | 1, 2, 3, 4, 5, 6 | Vertical well system |
| Asai et al. [38] | Injection and artificial fractures design | Single objective | Polynomial regression | Offline | N/A | 5 | Doublet well system |
| Samin et al. [40] | Injection-production and well-placement design | Multi-objective | N/A | N/A | GA | 5 | Vertical well system |
| Pollack and Mukerji [5] | Injection-production and well- | Single objective | N/A | N/A | PSO | 8 | Horizontal well system |
| | placement design | | | | | | |
| Song et al. [2] | Injection-production design | Multi-objective | N/A | N/A | NSGA-II | 3 | Multilateral-well system |
| Wang et al. [41] | Injection-production design | Single objective | LSTM | Offline | DE | 225 | Vertical well system |
| Wang et al. [42] | Well-placement design | Single objective | Random forest | Offline | GA | 2 | Doublet well system |
| This work | Injection-production design | Single objective | RBF, PNN | Online | DE | 80, 100, 160 | Vertical well system |

Solving high-dimensional expensive optimization problems has been a long-lasting concern in the areas of intelligent computations, geo-technique, hydrogeology, and energy exploitation [5, 29, 35, 43]. To tackle the challenge, researchers utilized global and local surrogate strategy [44] and ensemble surrogate strategy [45] to balance the exploration and exploitation. Liu et al. [46] proposed a Gaussian process surrogate model assisted evolutionary algorithm incorporating Sammom mapping to reduce the dimension for medium-scale computationally expensive optimization problems. Sun et al. [47] developed a surrogate-assisted cooperative swarm optimization algorithm with a surrogate-assisted particle swarm optimization and a surrogate-assisted social learning-based particle swarm optimization to cooperatively search for the global optimum. Zhen et al. [48] presented a two-stage data-driven evolutionary optimization to achieve early exploration and later exploitation during the optimization process. Chen et al. [39] proposed an efficient hierarchical surrogate-assisted differential evolution algorithm towards high-dimensional expensive optimization problems to alleviate the curse of dimensionality. Wang et al. [43] proposed an evolutionary sampling assisted optimization method, in which a global RBF model was employed to choose the best offspring and built local surrogate to accelerate search. Wei et al. [49] employed gradient boosting classifier as the surrogate model to select promising candidates for real function evaluation. Liu et al. [50] developed a surrogate-assisted multi-population particle swarm optimizer using affinity propagation clustering to generate several subswarms and guide the search of each subswarm incorporating a diversity maintenance scheme. Most of the aforementioned algorithms perform well on low-dimensional optimization problems. Besides, most existing SAEAs adopt regression models as surrogate to guide the optimization process. Classifier as the surrogate to guide the optimization process is still in its infancy and needs further exploration.

This research aims to develop a novel surrogate-assisted evolutionary algorithm with accelerated convergence to solve the effective heat extraction problems of enhanced geothermal system. The proposed algorithm, namely surrogate-assisted level-based learning evolutionary search (SLLES), can be divided into classifier-assisted level-based learning pre-screen part and local evolutionary search part. Concretely, probabilistic neural network (PNN) is employed as classifier to classify the offspring into pre-set number of levels. Level-based learning strategy is developed to generate much promising and informative candidates pre-screened by classifier and RBF, and then real simulation evaluation is conducted. Subsequently, for local evolutionary search part, a surrogate model is constructed at the local promising area. The optimum of the surrogate model obtained by the optimizer is selected to conduct real simulation evaluation. The cooperation of level-based learning strategy and local evolutionary search strategy is expected to balance the exploration and exploitation during the optimization process. After iteratively sampling from the design space, the robustness and effectiveness of the algorithm is believed to be improved significantly. The proposed algorithm is capable of taking advantages of both the level-based learning strategy and the classification model to improve the robustness and scalability of SAEAs in dealing with high-dimensional expensive optimization problems.

The main works of this study is summarized as: (1) A novel algorithm SLLES is proposed in solving heat extraction optimization problems for EGS. SLLES is divided into classifier-assisted level-based learning pre-screen part and local evolutionary search part. (2) A new surrogate model management strategy, i.e., classifier-assisted level-based learning pre-screen strategy, is developed to enhance the search ability of the algorithm. (3) Two synthetic EGS cases are conducted to test the performance of the proposed algorithm. As a general data-driven optimization framework, the developed workflow intends to shed light on the model management strategies of optimization in the areas of energy exploitation for complex system involving time-consuming simulation prediction, and shows potential applications in other energy systems, e.g., wind turbine performance optimization [51], solar photovoltaic energy optimization [52, 53], hydrocarbon resources production optimization [44] and $CO_2$ sequestration storage [54, 55].

## 2. Optimization design for heat extraction of EGS

### 2.1 Forward hydrothermal simulation

The mass conservation and momentum equations of subsurface groundwater flow in porous medium for geothermal reservoir are expressed as follows [7]:

$$\frac{\partial}{\partial t}(\rho_f \phi_m) + \nabla(\rho_f u_m) = \Psi_{mf} + Q_m \quad (1)$$

$$u_m = -\frac{k_m}{\mu}(\nabla p + \rho_f g \nabla D) \quad (2)$$

where $t$ is the time (s), $\rho_f$ is the density of subsurface fluid ($kg/m^3$), $\phi_m$ is the porosity of matrix, $u_m$ is the velocity of fluid ($m/s$), $\Psi_{mf}$ is the flux transfer function between fractures and matrix $kg/(m^3 \cdot s)$, $Q_m$ is the mass source/sink term ($kg/(m^3 \cdot s)$), $k_m$ is the matrix permeability ($m^2$), $\mu$ is the viscosity of fluid ($Pa \cdot s$), $p$ is the pressure ($Pa$), $g$ is gravitational acceleration ($m/s$), and $\nabla D$ is a unit vector along the direction of gravity ($m$). The flux transfer function $\Psi_{mf}$ is determined by:

$$\Psi_{mf} = \nabla \cdot \left(\rho \cdot \left(-\frac{k_m}{\mu_f}\nabla P_{boundary}\right)\right) \quad (3)$$

where $\nabla P_{boundary}$ denotes the pressure gradient performed in the tangential plane of a fracture. The mass conservation equation in fractures is expressed as:

$$\frac{\partial}{\partial t}(\rho \phi_f) + \nabla_T(\rho u_f) = \Psi_{fm} + Q_m \quad (4)$$

where $\phi_f$ is the porosity of fractures, $u_f$ is the velocity of fluid in fractures $m/s$, $\nabla_T$ is the gradient restricted to the tangential plane of the fracture. Fracture flow is considered to obey Darcy's law:

$$u_f = -\frac{k_f}{\mu}(\nabla_T p + \rho g \nabla_T D) \quad (5)$$

where $u_f$ is the fluid velocity in fractures ($m/s$), $k_f$ is the permeability of fracture ($m^2$) which is calculated using the cubic law $k_f = b^2/12$, $b$ is the aperture of fracture (m), and $\Psi_{fm}$ is the flux transfer function from matrix to fractures $kg/(m^3 \cdot s)$.

The heat transfer in matrix of EGS follows the energy conservation equation, which can be expressed as:

$$(\rho C)_{eff}\frac{\partial T}{\partial t} + \rho_f C_f u_m \cdot \nabla T - \nabla \cdot (\lambda_{eff}\nabla T) = E_{mf} + Q_{hm} \quad (6)$$

where $C$ is the specific heat capacity ($J/(kg \cdot K)$), $(\rho C)_{eff} = \phi \rho_f C_f + (1-\phi)\rho_s C_s$ denotes the effective heat capacity ($J/(kg \cdot K)$), $\lambda_{eff} = \phi \lambda_f + (1-\phi)\lambda_s$ represents the effective thermal conductivity ($W/(m \cdot K)$), $\lambda$ is the heat conductivity ($W/(m \cdot K)$) with subscripts $s$ and $f$ representing the formation solid and fluid respectively, $E_{mf} = h \cdot (T_f - T_m)$ is the heat transfer from fractures to matrix ($W/(m^3)$), $h$ is the convective heat transfer coefficient ($W/(m^2 \cdot K)$) between matrix and fractures, and $Q_{hm}$ is the heat source/sink term in matrix ($W/(m^3)$).

The energy conservation equation in fractures of EGS is expressed as:

$$b\rho_f C_f \frac{\partial T}{\partial t} + b\rho_f u_f C_f \nabla_T T - \nabla_T \cdot (b\lambda_f \nabla_T T) = E_{fm} + Q_{hf} \tag{7}$$

where $E_{fm} = h \cdot (T_m - T_f)$ is the heat transfer from matrix to fractures ($W/(m^3)$), and $Q_{hf}$ is the heat source/sink term in fractures ($W/(m^3)$).

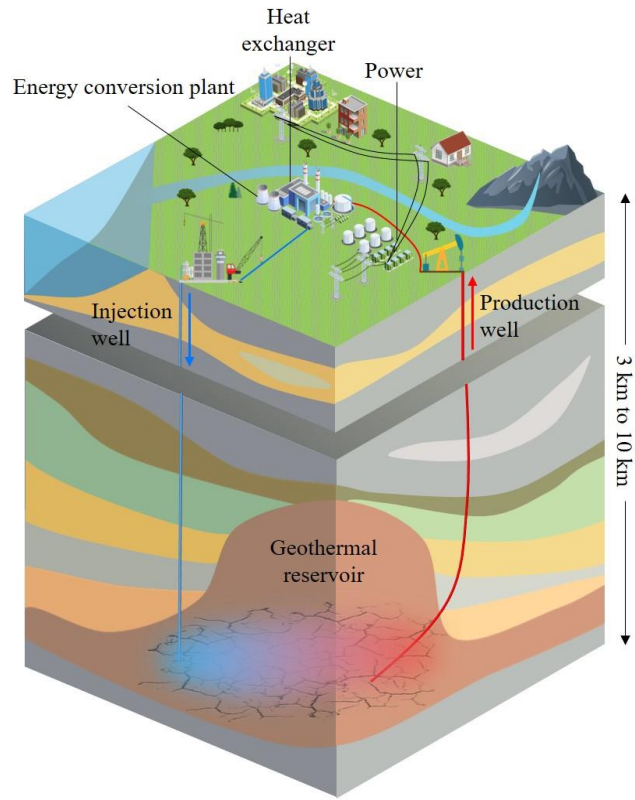

**Fig. 1.** Schematic diagram of the enhanced geothermal system

**2.2 Objective function for injection-production design**

As shown in **Fig. 1**, an EGS, consisting of complex above-ground and underground facilities, injects water into high-temperature geothermal reservoir to re-open the natural fractures and extract heated water for electricity generation. The heat extraction performance of EGS is evaluated by economic calculation for the output of forward hydrothermal simulation. The objective for heat extraction optimization problems is to maximize the net present value (NPV) of the geothermal reservoir during the project life. Since water is selected as the working liquid in this work, the profit for developing the geothermal reservoir represents the net energy production value minus the cost of water production and injection:

$$NPV(x,z) = CTEP \times r_e - CWI \times r_i - CWP \times r_p \tag{8}$$

where $x$ is the well-control variables, $z$ is the state variables calculated from hydro-thermal numerical simulation (e.g., temperature and pressure), $CTEP$, $CWI$, and $CWP$ denote the cumulative thermal energy production ($J$), the cumulative water injection ($m^3$), and the cumulative water production ($m^3$) of the geothermal reservoir, respectively, $r_e$, $r_i$ and $r_p$ denote the price of thermal energy per watt-hour ($\$/J$), the cost of water injection ($\$/(m^3)$) and the cost of water production ($\$/(m^3)$). Thus, the optimization problem can be summarized as:

$$\max NPV(x,z), \ x \in \mathbb{R}^d \tag{9}$$

s.t.

$$lb \leqslant x \leqslant ub \tag{10}$$

$$c(x) \leqslant 0 \tag{11}$$

where $lb$ and $ub$ are lower boundary and upper boundary of decision variables, $c(x)$ is the constraints of decision variables (e.g., physical and operational constraints).

## 3. Related optimization techniques

### 3.1 Differential evolution

Differential evolution (DE), firstly proposed by Storn and Price [56], is a traditional population-based metaheuristic algorithm for global optimization. Due to the simplicity of structure, DE is chosen as the optimizer for the proposed surrogate-guided evolutionary framework. DE consists of 4 parts: initialization, mutation, crossover, and selection. The detailed procedure is as follows: Initialize $NP$ individuals from the design space $[lb, ub]$ to generate the initial population $P = [x_1, ..., x_i, ..., x_{NP}], i = \{1, ..., NP\}, x_i \in \mathbb{R}^d$. Generate the mutant vectors $[v_1, ..., v_i, ..., v_{NP}], i = \{1, ..., NP\}, v_i \in \mathbb{R}^d$ with mutation operator. Commonly used mutation operators are as follows:

DE/rand/1

$$v_i = x_{r_1} + Mu \times (x_{r_2} - x_{r_3}) \tag{12}$$

DE/current-to-rand/1

$$v_i = x_i + Mu \times (x_{r_1} - x_i) + Mu \times (x_{r_2} - x_{r_3}) \tag{13}$$

DE/best/1

$$v_i = x_{best} + Mu \times (x_{r_1} - x_{r_2}) \tag{14}$$

DE/current-to-best/1

$$v_i = x_i + Mu \times (x_{best} - x_i) + Mu \times (x_{r_1} - x_{r_2}) \tag{15}$$

where $r_1$, $r_2$ and $r_3$ are randomly generated integers from 1 to $NP$, $Mu$ is the mutation factor, and $x_{best}$ is the best solution in the population. Subsequently, trial vectors are generated according to crossover operator, which can be expressed as:

$$u_i^j = \begin{cases} v_i^j & if \ rand \leqslant CR \ or \ j = j_{rand} \\ x_i^j & otherwise \end{cases}, j \in \{1,...,d\} \tag{16}$$

where $u_i^j$ is the $j^{th}$ dimension of the $i^{th}$ trial vector, $rand$ is a random number from 0 to 1, and $CR$ is the crossover factor. Then evaluate the function value of trial vectors $u$ and select new individuals for the next generation:

$$x_i' = \begin{cases} u_i & if \ f(u_i) < f(x_i) \\ x_i & otherwise \end{cases} \tag{17}$$

where $x_i'$ is the $i^{th}$ individual of the next generation.

### 3.2 Radial basis function

RBF has been widely applied to many scientific and engineering problems due to the promising ability to approximate the landscape of the objective function with small number of sample points [33, 47]. Since the approach is not sensitive to the dimension and also exhibits good approximation performance for high-dimensional problems, RBF is selected as surrogate in this work. RBF is an interpolation model consisting of a weighted sum of basis functions:

$$\hat{f}(x) = \sum_{i=1}^{n} \omega_i \varphi(\|x - c_i\|) \tag{18}$$

where $\omega$ is the weight coefficient, $\varphi(\cdot)$ is the basis function, $\|\cdot\|$ is the Euclidean norm, and $c$ is the centre points of the basis function. Specifically, for given n training sample points $x = [x_1,...,x_n]$, with function value $y = [y_1,...,y_n]$, then $\omega$ can be obtained as:

$$\omega = \varphi^{-1} y \tag{19}$$

**3.3 Probabilistic neural network**

Probabilistic neural network (PNN), a feed forward neural network, consists of an input layer, a pattern layer, a summation layer and an output layer. The number of input neurons equals the dimension of decision variables. The pattern layer is able to calculate the probability density estimation from the input sample points to the kernel centre as follows:

$$\varphi_{ij}(x) = \frac{1}{(2\pi)^{d/2} \sigma^d} \exp\left[ -\frac{(x-c_{ij})^T (x-c_{ij})}{2\sigma^2} \right] \tag{20}$$

where $\sigma$ is the kernel width, and $c_{ij}$ is the $j^{th}$ kernel centre vector for the $i^{th}$ class. Afterwards, the summation layer calculates the maximum likelihood of x being classified into the $i^{th}$ class by:

$$p_i(x) = \frac{1}{N_i} \sum_{j=1}^{N_i} \varphi_{ij}(x) \tag{21}$$

where $N_i$ is the number of training sample points belonging to the $i^{th}$ class. The output layer selects the maximum $p_i(x)$ as the final classification result.

**4 The proposed SLLES algorithm**

To solve the simulation-involved optimization problems more efficiently, a novel algorithm SLLES is proposed for heat extraction optimization of EGS. SLLES contains classifier-assisted level-based learning pre-screen part and local evolutionary search part. The detailed information of SLLES will be introduced in this section.

**4.1 Classifier-assisted level-based learning pre-screen sampling**

The framework of classifier-assisted level-based learning pre-screen sampling is presented in **Fig. 2**. The pseudo-code is shown in **Algorithm 1**. The sample points are initialized by Latin Hypercubic Sampling (LHS), and objective function values are evaluated with numerical simulation. The initial population is generated by selecting $NP$ most promising sample points and sorted into $k$ levels according to the ranking of objective values with each level containing $NP/k$ individuals. Then the classifier is trained with current population to predict the level of newly generated offspring. Afterwards, offspring is generated by the proposed classifier-assisted level-based learning mutation operator, the variant of DE mutation operator:

$$v_i = x_i + Mu \times (x_{r_1} - x_i) + rand \times Mu \times (x_{r_2} - x_i) \qquad (22)$$

where $x_{r_1}$ is an individual randomly selected from the first level, $x_{r_2}$ is an individual randomly selected from the first and second level. When $x_i$ is from the first or the second level, $x_{r_1}$ and $x_{r_2}$ are both randomly selected from the first level. Then use conventional DE crossover operation to generate new offspring. After that, the classifier is used to identify the promising individuals predicted in the first level. The level-based learning mutation and prediction of classifier keeps until $N_f$ of the population are predicted to be the first level. Subsequently, the most promising individual is identified using RBF with the predicted function value:

$$\hat{u}_{best} = \arg\min_{u_i \in \{u\}^1} \hat{f}(u_i) \qquad (23)$$

where $\hat{u}_{best}$ is the most promising solution pre-screened by RBF $\hat{f}$, and $\{u\}^1$ is the selected trial vector set as the first level. After evaluating the objective function value, the sample point is added into database $D$. If the maximum simulation evaluation is achieved, the algorithm is terminated and the optimization result will be output. Otherwise, the optimization loop will be continued.

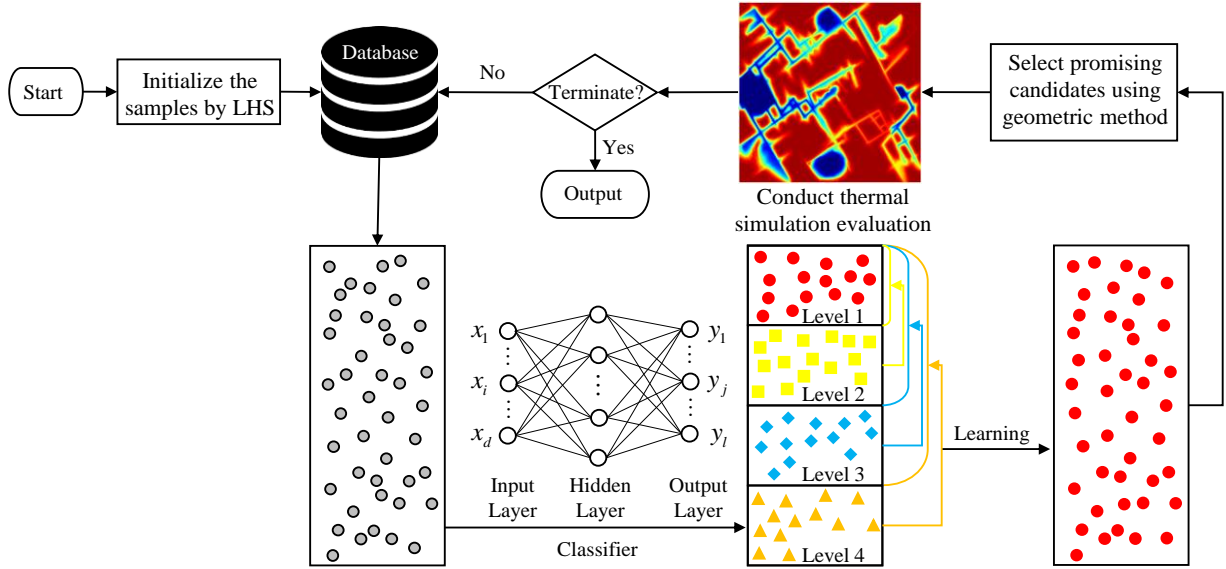

**Fig. 2.** The framework of classifier-assisted level-based learning pre-screen sampling

**Algorithm 1** Pseudo-code of classifier-assisted level-based learning pre-screen sampling

**Input**: Database $D$, the population size $NP$, the number of levels $k$, mutation factor $Mu$, crossover factor $CR$, the maximum number of simulation runs $FEs_{\max}$

1:    Generate $NP$ initial sample points by LHS;

2:    Evaluate the initial sample points with hydro-thermal simulation;

3:    $FEs = NP$;

4:    **While** $FEs < FEs_{\max}$

5:        Select $NP$ best sample points $P = \{x_1, x_2, ..., x_{NP}\}$ from database $D$;

6:        Sort the population $P$ into $k$ levels, with each level containing $NP/k$ individuals;

7:        Use PNN as the classifier with the level-based population;

8:        Use classifier-assisted level-based learning mutation operator (Eq. 22) to generate $NP$ new solutions $u = \{u_1, u_2, ..., u_{NP}\}$;

9:        Predict the level of the offspring using PNN;

10:       **While** the number of offspring predicted in the first level $< N_f$

11:           **For** each candidate in $u$ that is not predicted in the first level

12:               Evolve the solution with classifier-assisted level-based learning mutation operator;

13:               Predict the level of the solution using PNN;

14:           **End for**

15:       **End while**

16:       Train RBF surrogate with all sample points in database $D$;

17:       Identify the most promising sample point predicted in the first level $\hat{u}_{best}$ with RBF;

18:       Evaluate the objective function value of $\hat{u}_{best}$ with simulator;

19:       Add the new sample point and corresponding function value into database $D$;

20:       $FEs = FEs + 1$;

21:   **End while**

**Output**: The best solution $x_{best}$ and database $D$.

## 4.2 Local evolutionary search sampling

To further exploit the promising regions near the currently found optima and accelerate the convergence of the proposed algorithm, local evolutionary search sampling is employed. **Fig. 3** presents the framework of local evolutionary search sampling, and **Algorithm 2** shows the pseudo code of local evolutionary search sampling. Specifically, $\tau$ newest solutions are selected from database $D$ as the training sample points. After that, the upper and lower bound of the local reduced region are calculated as:

$$\begin{cases} lb_l = \min(x_1, x_2, ..., x_\tau) \\ ub_l = \max(x_1, x_2, ..., x_\tau) \end{cases} \quad (24)$$

where $lb_l$ and $ub_l$ are the lower and upper bound of the local surrogate $\hat{f}_l$, respectively. Subsequently, local surrogate $\hat{f}_l$ is built in the local reduced region, and the optimum $\hat{x}_{best}$ of the local surrogate is located by DE optimizer based on the criterion:

$$\hat{x}_{best} = arg\min \hat{f}_l(x) \quad (25)$$

After that, evaluate the objective function value of $\hat{x}_{best}$ with simulator, and add the new sample point and corresponding function value into database $D$. The optimization loop continues until the maximum simulation evaluation is reached.

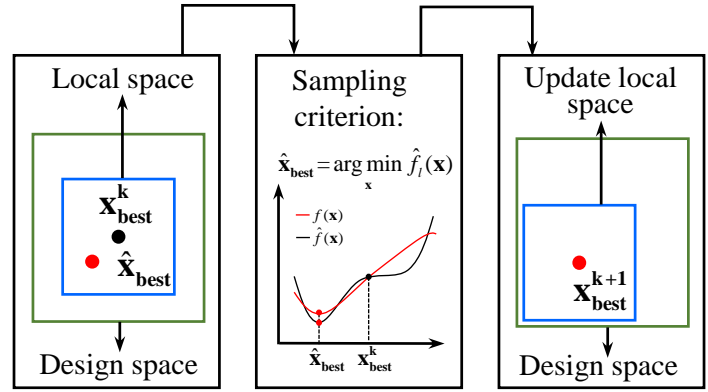

**Fig. 3.** The framework of local evolutionary search sampling

---

**Algorithm 2** Pseudo code of local evolutionary search sampling

---

**Input**: Database $D$, size of training sample points $\tau$, mutation factor $Mu$, crossover factor $CR$, the maximum number of simulation runs $FEs_{\max}$

1:    Generate $\tau$ initial sample points by LHS;
2:    Evaluate the initial sample points with hydro-thermal simulation;
3:    $FEs = \tau$;
4:    **While** $FEs < FEs_{\max}$
5:       Select $\tau$ newest solutions from database $D$ as the training sample points;
6:       Calculate the upper and lower bound $[lb_l, ub_l]$ of the local reduced region;

7:     Build a local surrogate $\hat{f}_l$ from the local reduced region;

8:     Find the optimum $\hat{x}_{best}$ of the local surrogate $\hat{f}_l$ with DE optimizer;

9:     Evaluate the objective function value of $\hat{x}_{best}$ with simulator;

10:    Add the new sample point and corresponding function value into database $D$;

11:    $FEs = FEs + 1$;

12: **End while**

**Output**: The best solution $x_{best}$ and database $D$.

### 4.3 The framework of SLLES

SLLES consists of classifier-assisted level-based learning pre-screen sampling and local evolutionary search sampling. The workflow of the proposed SLLES algorithm is shown in **Fig. 4** and the pseudo-code of SLLES is presented in **Algorithm 3**. Specifically, the classifier-assisted level-based learning strategy employs PNN as classifier to classify the offspring into pre-set number of levels. The offspring in different levels use level-based learning strategy to generate more promising and informative candidates pre-screened by classifier to conduct real simulation evaluations. In the local evolutionary search part, surrogate model is constructed at the local promising area. The optimum of the surrogate model obtained by the optimizer is selected to conduct real simulation evaluations. Classifier-assisted level-based learning pre-screen sampling is able to explore promising but relatively uncertain area, while local evolutionary search sampling is able to further exploit the area near the current optima and speed up the convergence. The cooperation of the two parts can balance the exploration and exploitation during the optimization process. After iteratively sampling from the design space, the robustness and effectiveness of the algorithm can be improved significantly.

**Algorithm 3** Pseudo-code of SLLES

**Input**: Database $D$, the population size $NP$, the number of levels $k$, mutation factor $Mu$, crossover factor $CR$, size of training sample points $\tau$, the maximum number of simulation runs $FEs_{\max}$

1:    Generate $NP$ initial sample points by LHS;

2:    Evaluate the initial sample points with hydro-thermal simulation;

3:    $FEs = NP$;

4:    **While** $FEs < FEs_{\max}$
5:       // Classifier-assisted level-based learning pre-screen sampling
6:       Select $NP$ best sample points $P = \{x_1, x_2, ..., x_{NP}\}$ from database $D$;
7:       Sort the population $P$ into $k$ levels, with each level containing $NP/k$ individuals;
8:       Use PNN as the classifier with the level-based population;
9:       Use classifier-assisted level-based learning mutation operator (Eq. 22) to generate $NP$ new solutions $u = \{u_1, u_2, ..., u_{NP}\}$;
10:      Predict the level of the offspring using PNN;
11:      **While** the number of offspring predicted in the first level $< N_f$
12:         **For** each candidate in $u$ that is not predicted in the first level
13:           Evolve the solution with classifier-assisted level-based learning mutation operator;
14:           Predict the level of the solution using PNN;
15:         **End for**
16:      **End while**
17:      Train RBF surrogate with all sample points in database $D$;
18:      Identify the most promising sample point predicted in the first level $\hat{u}_{best}$ with RBF;
19:      Evaluate the objective function value of $\hat{u}_{best}$ with simulator;
20:      Add the new sample point and corresponding function value into database $D$;
21:      $FEs = FEs + 1$;
        // Local evolutionary search sampling
22:      Select $\tau$ newest solutions from database $D$ as the training sample points;
23:      Calculate the upper and lower bound $[lb_l, ub_l]$ of the local reduced region;
24:      Build a local surrogate $\hat{f}_l$ from the local reduced region;
25:      Find the optimum $\hat{x}_{best}$ of the local surrogate $\hat{f}_l$ with DE optimizer;
26:      Evaluate the objective function value of $\hat{x}_{best}$ with simulator;
27:      Add the new sample point and corresponding function value into database $D$;
28:      $FEs = FEs + 1$;
29: **End while**

**Output**: The best solution $x_{best}$ and database $D$.

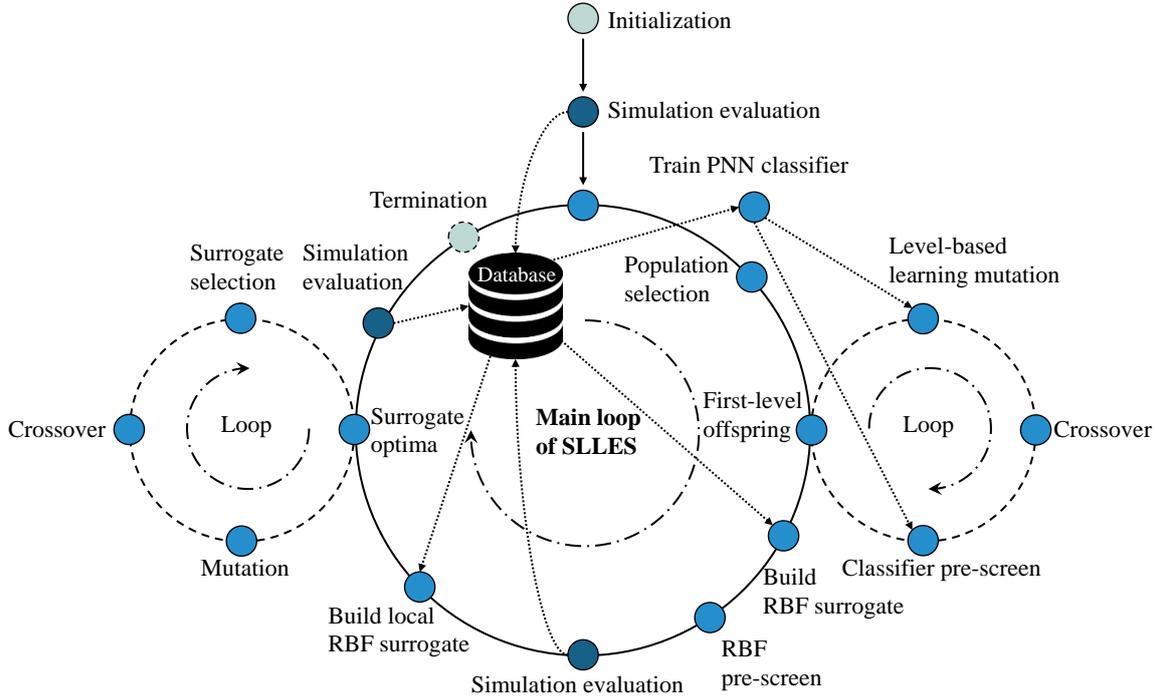

**Fig. 4.** The workflow of the proposed SLLES algorithm

## 5. Case studies

In this section, experiments on three cases are conducted to validate the performance of SLLES. Case 1 tests four 100D benchmark functions in comparison with other state-of-the-art algorithms on high-dimensional expensive optimization problems. To further examine the effectiveness of SLLES on heat extraction optimization of EGS, comparative experiments on a 2D theoretical fractured reservoir and a 3D field-scale EGS are conducted.

**5.1 Case 1: Experiment on four 100D benchmark functions**

To verify the effectiveness of the proposed algorithm, four 100D benchmark function problems with different properties are selected to locate the minimum of the problems in pre-set number of function evaluations. The description for the properties of four 100D benchmark functions is summarized in **Table 2**, and the 3D map for the 2D fitness landscape of 4 benchmark functions is shown in **Fig. 5**. The detailed expression of the benchmark functions can be found in Suganthan et al. [57]. The stopping criterion for each problem is 1000 maximum number of real function evaluations.

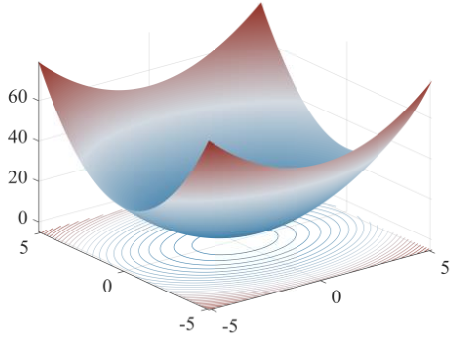
(a) Ellipsoid function

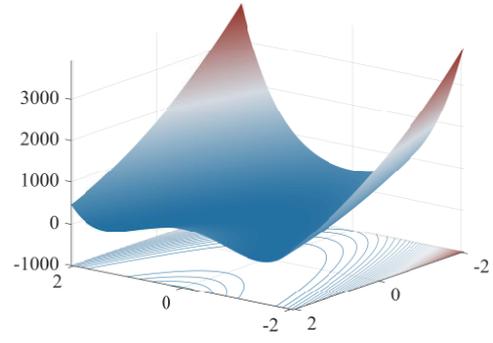
(b) Rosenbrock function

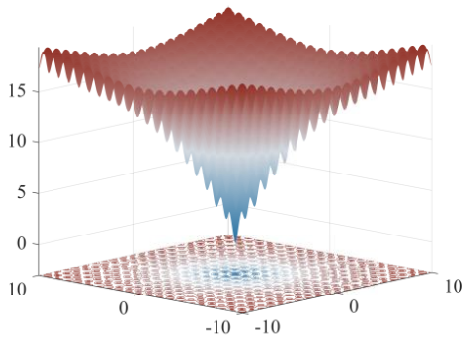
(c) Ackley function

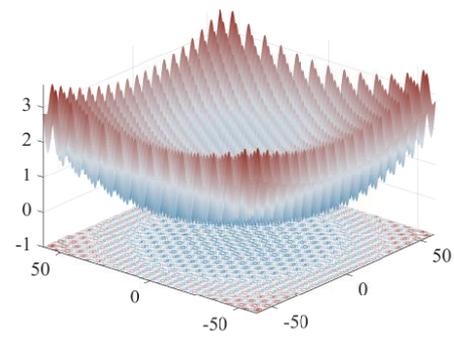
(d) Griewank function

**Fig. 5.** 3D map for 2D fitness landscape of 4 benchmark functions: (a) Ellipsoid function; (b) Rosenbrock function; (c) Ackley function; and (d) Griewank function

**Table 2.** Properties of the benchmark functions used in the experiment

| Benchmark name | D | Design space | Global optimum | Characteristics |
| --- | --- | --- | --- | --- |
| Ellipsoid (F1) | 100 | $[-5.12, 5.12]^D$ | 0 | Unimodal |
| Rosenbrock (F2) | 100 | $[-2.048, 2.048]^D$ | 0 | Multimodal |
| Ackley (F3) | 100 | $[-32.768, 32.768]^D$ | 0 | Multimodal |
| Griewank (F4) | 100 | $[-600, 600]^D$ | 0 | Multimodal |

To investigate the contribution of each sampling strategy to the proposed algorithm framework, experiments are conducted using DE, classifier-assisted level-based learning pre-screen sampling (CLLPS), local search sampling (LS) and SLLES. **Table 3** presents the results of different sampling strategies on the benchmark problems, and **Fig. 6** presents corresponding convergence curve during the optimization process. As shown in **Fig. 6**, all surrogate-assisted algorithms outperform DE significantly because the surrogate provides informative predictions for the objective function. Although holding great global exploration ability, CLLPS converges slowly in the early stage since it spends a large number of computational resources on exploring relatively uncertain area. LS converges efficiently in the early optimization stage, but is prone to suffering from local optima in the late stage. LS exhibits promising performance on F1 because Ellipsoid function is a unimodal problem and LS is less likely to get stuck in a local optimum. The results demonstrate that local search sampling can speed up the convergence in the early stage, while classifier-assisted level-based learning pre-screen sampling can enhance the exploration ability to alleviate trapping into local optima in the late stage. The cooperation of the two sampling strategies can balance the exploration and exploitation during the optimization process. After iteratively sampling from the design space, the robustness and effectiveness of the algorithm can be improved significantly.

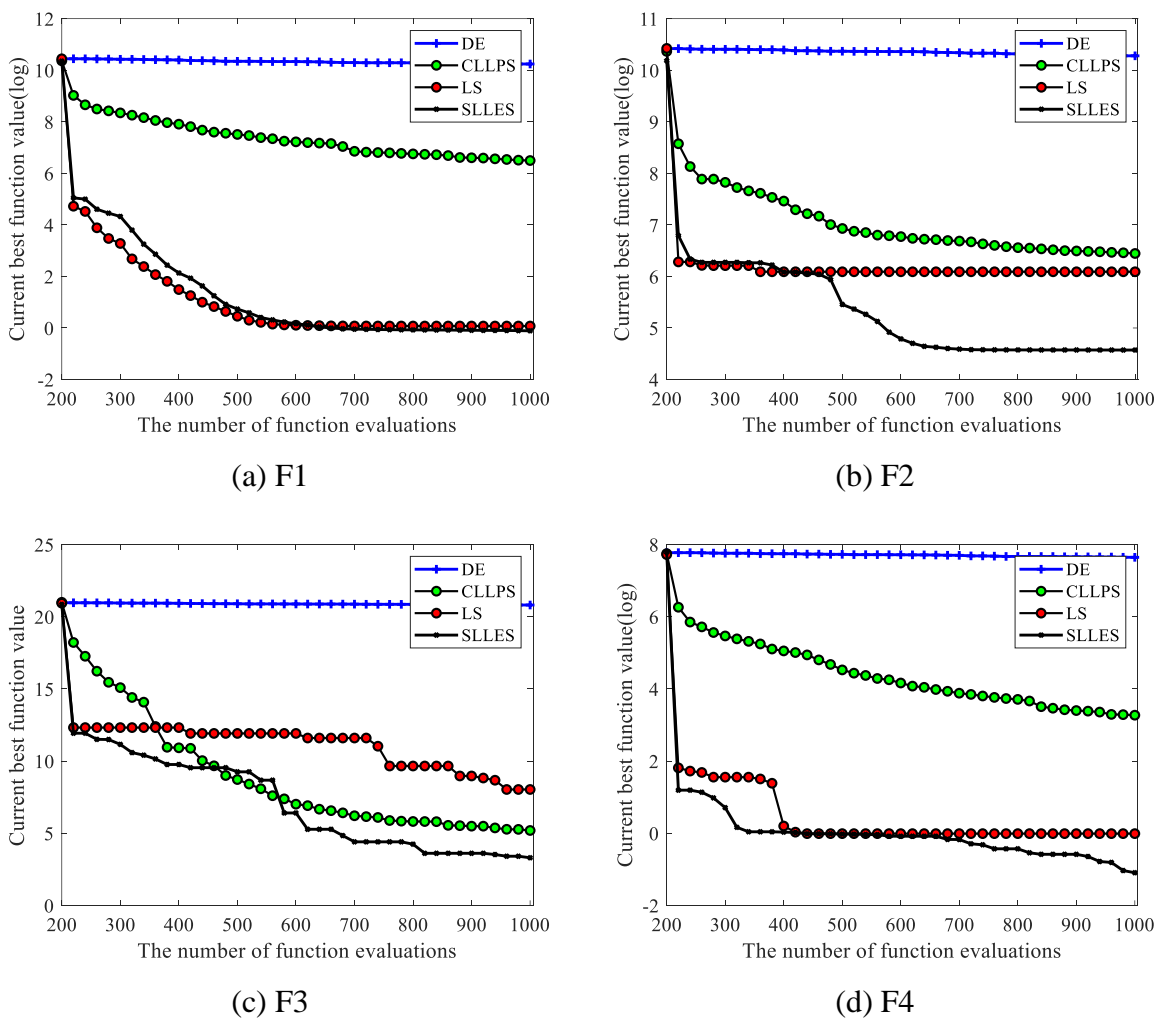

(a) F1    (b) F2    (c) F3    (d) F4

**Fig. 6.** Convergence curves of DE, CLLPS, LS and SLLES on 100D benchmark problems: (a) F1; (b) F2; (c) F3; and (d) F4

**Table 3.** Results of different sampling strategies on the problems

| Problem | DE | LS | CLLPS | SLLES |
|---|---|---|---|---|
| F1 | 1.0332E+03 | 7.6106E+01 | 6.6287E+02 | **8.9612E-01** |
| F2 | 2.7142E+03 | 1.6559E+02 | 6.2972E+02 | **9.6748E+01** |
| F3 | 1.5756E+01 | 4.1134E+00 | 5.1910E+01 | **3.3005E+00** |
| F4 | 6.3353E+01 | 9.8898E-01 | 2.6341E+01 | **3.3459E-01** |

Five state-of-the-art surrogate-assisted evolutionary algorithms, namely, SA-COSO [47], SHPSO [33], ESAO [43], CA-LLSO [49] and SA-MPSO [50], are selected to compare the performance of the proposed algorithm on benchmark functions. The optimization results of the comparison algorithms are adopted from the corresponding papers. Results of SA-COSO, SHPSO, ESAO, SA-MPSO and SLLES on the benchmark problems are summarized in **Table 4**, and corresponding convergence curves are presented in **Fig. 7**. Since CA-LLSO didn't provide the convergence curve, only final optimization result after 1000 function evaluations is compared in **Table 4**. Overall, SLLES outperform the five state-of-the-art algorithms on the four 100D benchmark problems (**Table 4** and **Fig. 7**). SA-MPSO and SHPSO show good and similar performance on the problems. The optimization results of benchmark problems can confirm that SLLES is competitive and holds state-of-the-art simulation-involved optimization framework.

**Table 4.** Results of SA-COSO, SHPSO, ESAO, SA-MPSO and SLLES on all the problems.

| Problem | SA-COSO | SHPSO | ESAO | CA-LLSO | SA-MPSO | SLLES |
|---|---|---|---|---|---|---|
| F1 | 1.0332E+03 | 7.6106E+01 | 1.2829E+03 | 1.2839E+03 | 1.64E+01 | **8.9612E-01** |
| F2 | 2.7142E+03 | 1.6559E+02 | 5.7884E+02 | 5.4529E+02 | 1.48E+02 | **9.6748E+01** |
| F3 | 1.5756E+01 | 4.1134E+00 | 1.0364E+01 | 1.1614E+01 | 4.86E+00 | **3.3005E+00** |
| F4 | 6.3353E+01 | 1.0704E+00 | 5.7342E+01 | 1.0205E+02 | 9.40E-01 | **3.3459E-01** |

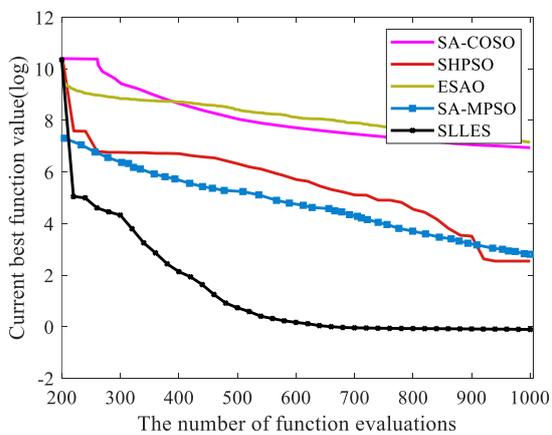
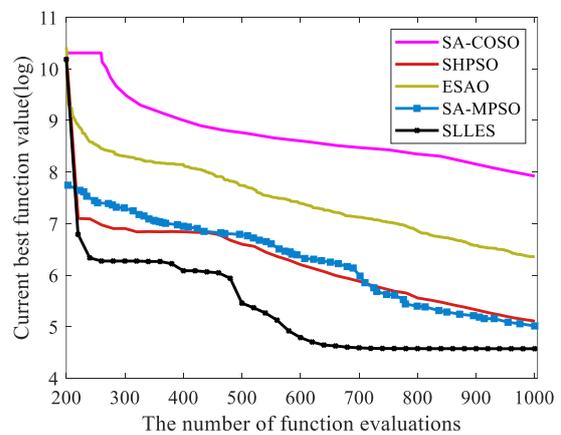

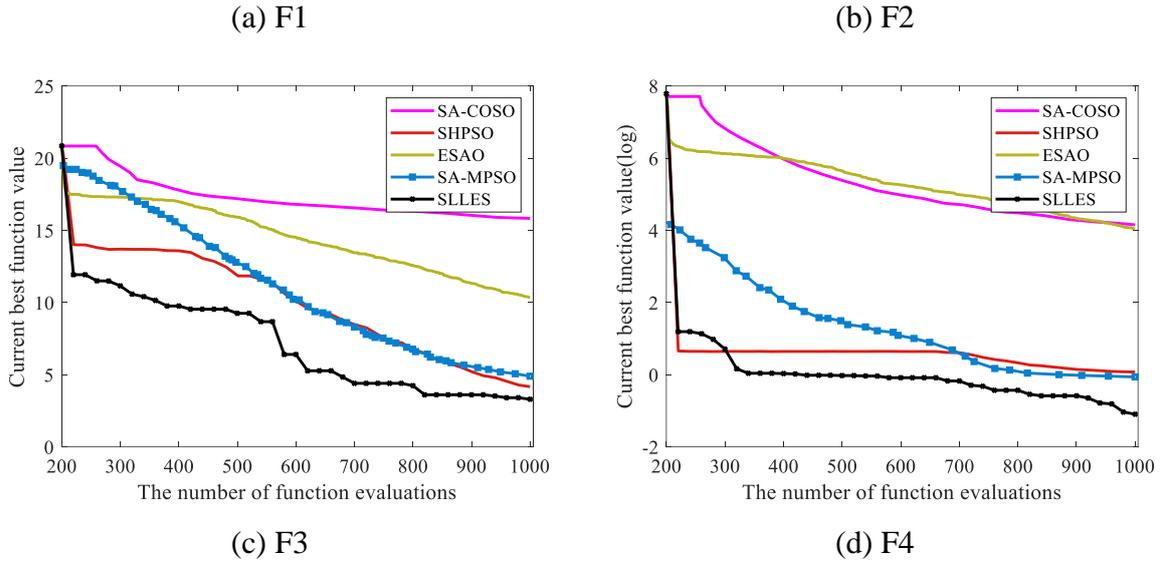

(a) F1            (b) F2

(c) F3            (d) F4

**Fig. 7.** Convergence curves of SA-COSO, SHPSO, ESAO, SA-MPSO and SLLES on 100D benchmark problems: (a) F1; (b) F2; (c) F3; and (d) F4

### 5.2 Case 2: Application on a 2D fractured geothermal reservoir

A 2D synthetic fractured geothermal reservoir is employed as the base model to evaluate the effectiveness of SLLES in heat extraction optimization. The fracture geometry, orientation and length distribution of the fractured geothermal reservoir is presented in **Fig. 8**. The model involves two nearly orthogonal fracture sets. Four injection wells and five production wells are drilled in the reservoir. The size of the model is $1000 \times 1000 m$, with a thickness of $20m$. The detailed parameter settings of the fractured geothermal reservoir are listed in **Table 5**. The profit of the thermal reservoir during the project period is considered as the criterion to evaluate the production performance of EGS, which is related to the net energy production value and the cost of water injection and production (Eq. 8). The initial temperature of the geothermal reservoir is 200 ℃ and the temperature of injection water is 50 ℃. The lifetime of the geothermal reservoir project is 6000 days and each timestep is 300 days. The production wells are operated at constant bottom hole pressure of 10 MPa. Therefore, the goal is to determine the control scheme of 4 injection wells at 20 time steps, totally 80 variables to be optimized to maximize the NPV of the EGS model. The ideal goal is to increase the heat sweep efficiency while avoiding the uneconomical development during the lifetime of the project.

**Table 5.** Parameter settings of the fractured geothermal reservoir

| Parameter | Value |
|---|---|
| Initial temperature | 200 ℃ |
| Initial pressure | 30 MPa |
| Temperature of injection water | 50 °C |
| BHP range of production wells | [10MPa, 30 MPa] |
| Flow rate range of injection wells | $[0, 50] \times 10^{-3} m^3/s$ |
| Reservoir depth | 4000m |
| Porosity of rock matrix | 0.1 |
| Porosity of fracture | 0.5 |
| Permeability of rock matrix | $5 \times 10^{-15} m^2$ |
| Permeability of fractures | $10^{-7} m^2$ |
| Reservoir thickness | 50 m |
| Thermal conductivity of rock matrix | 2 W/(m*K) |
| Thermal conductivity of water | 0.698 W/(m*°C) |
| Heat capacity of rock matrix | 850 J/(kg °C) |
| Heat capacity of water | 4200 J/(kg °C) |

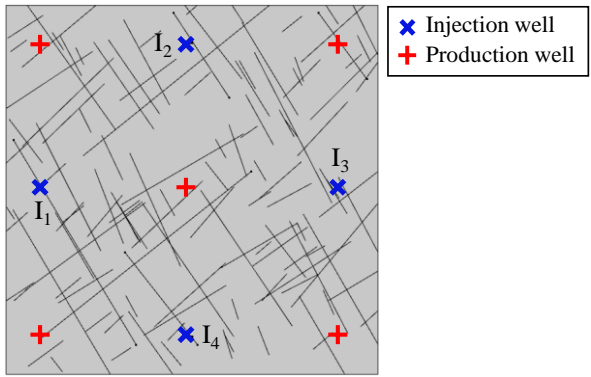
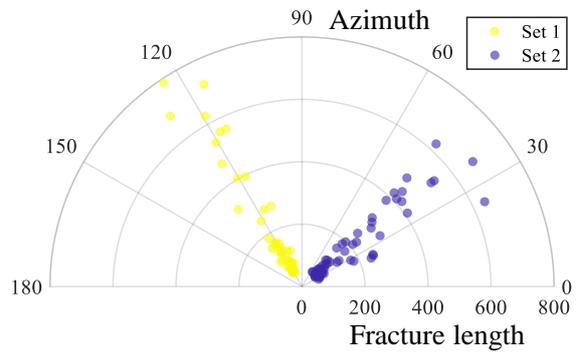

(a) DFN model  (b) Fracture orientation and length distribution

**Fig. 8.** Fracture geometry, orientation and length distribution of the 2D fractured geothermal reservoir: (a) DFN model; (b) Fracture orientation and length distribution

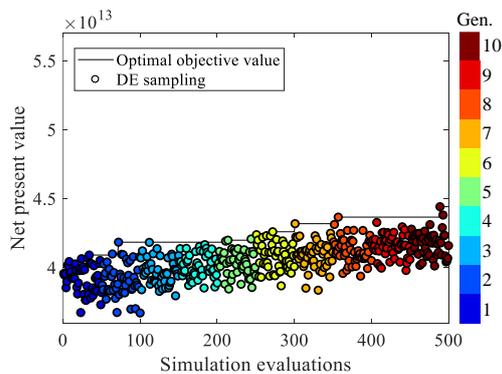
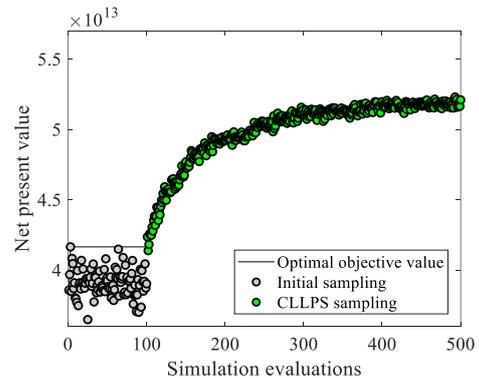

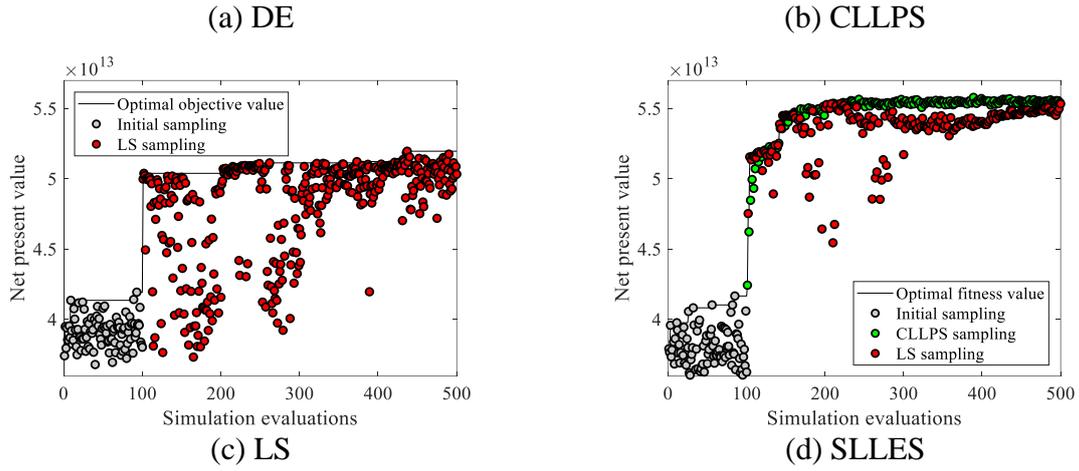

**Fig. 9.** Convergence curves and sampling processes of DE, CLLPS, LS and SLLES for the fractured geothermal reservoir.

The fractured geothermal model is optimized using DE, CLLPS, LS and SLLES algorithms. To construct the initial surrogate model, all surrogate-assisted algorithms generate 100 initial sample points into the database by LHS. After that, 400 online dynamic sampling are performed. That is, the total number of simulation evaluations is 500. **Fig. 9** illustrates the convergence curves and sampling process of different algorithms during the optimization. DE converges slowly in comparison with surrogate-assisted algorithms. CLLPS does not converge as fast as LS in the early optimization stage because it consumes large computational resource on exploring uncertain areas. After enough simulation evaluations, the well control scheme supplied by CLLPS becomes much better. LS shows efficient performance in the early optimization stage, but prone to trapping into local optima in the following iteration process. SLLES, combining the advantage of CLLPS and LS, converges fast in the early optimization stage and holds better exploration ability. After 500 simulation runs, SLLES obtains higher NPV.

**Fig. 10** shows the optimal well-control scheme supplied by DE, CLLPS, LS and SLLES for the fractured geothermal reservoir. Sorek et al. [58] stated that the optimal control trajectory can be always found at the boundary of several intervals, which means that the optimal well-control scheme always follows a bang-bang behaviour. The bang-bang solution or bang-singular solution will be operated as on-off of the well valves. All three surrogate-assisted algorithms supply bang-singular well-control schemes. Our previous work has demonstrated the effectiveness of interpolation control method in achieving smooth well-control trajectories [59]. The interpolation control method is a good choice if the project manager has a practical requirement to smooth the well-control trajectories. **Fig. 11** presents corresponding thermal distributions for the fractured geothermal reservoir after 1500, 3000 and 6000 days. Among all the algorithms, the final sweep efficiency of DE is relatively low, while the performances of three surrogate-assisted algorithms are comparative. The fracture density is higher in the lower left of the fracture model, resulting in high inter-well connectivity. Thus, the final optimized well-control scheme of SLLES injects less water for injection wells 1 and 4 at the early developing period, and injects more water for injection wells 2 and 3. Such production paradigm is able to improve the heat sweep efficiency of the fractured geothermal reservoir and maximize the NPV during the lifetime of the project. After 6000 days, most of the geothermal energy that can be displaced by the fracture network has been extracted, and most of the remaining untapped energy is distributed in the low-permeability matrix.

The geothermal reservoir development performances of four algorithms for case 2 are compared in detail considering average production temperature, cumulative thermal energy production, thermal energy production rate and cumulative NPV (**Fig. 12**). The average production temperature of SLLES decreases slowly and rise slightly between days 1500-2400 because the injection wells 1 and well 4 turn on and begin injecting water to sweep high-temperature water into production wells. The thermal energy production rates of all four methods generally maintain a decline trend. The cumulative thermal energy production and cumulative NPV maintain an increasing trend, while the growth rates decrease with time. The heat extraction performance of the optimal well-control scheme provided by DE is far worse than that by surrogate-assisted algorithms. SLLES achieves better heat extraction performance in comparison with DE and other surrogate-assisted optimization algorithms.

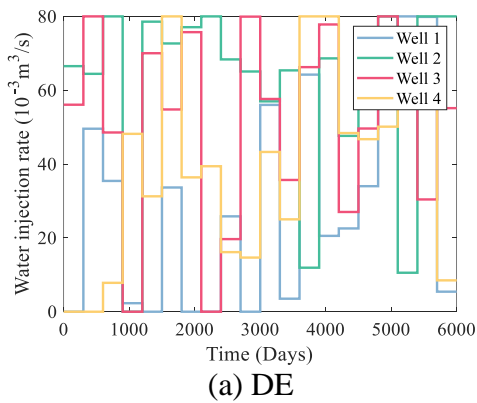
(a) DE

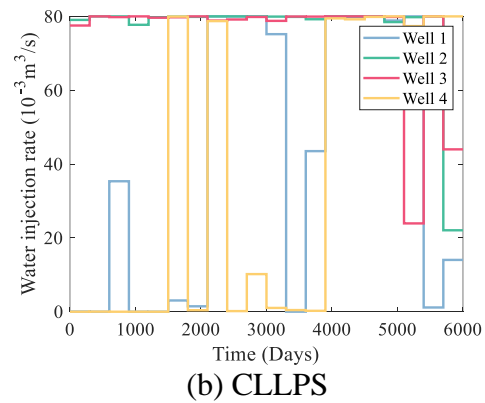
(b) CLLPS

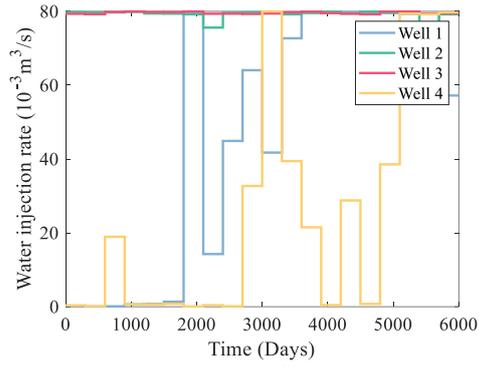
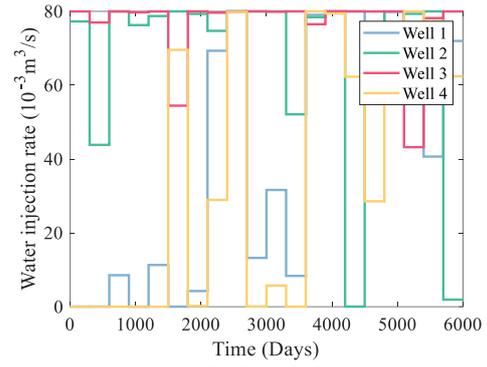

(c) LS  (d) SLLES

**Fig. 10.** The optimal well-control schemes obtained by DE, CLLPS, LS and SLLES for the fractured geothermal reservoir.

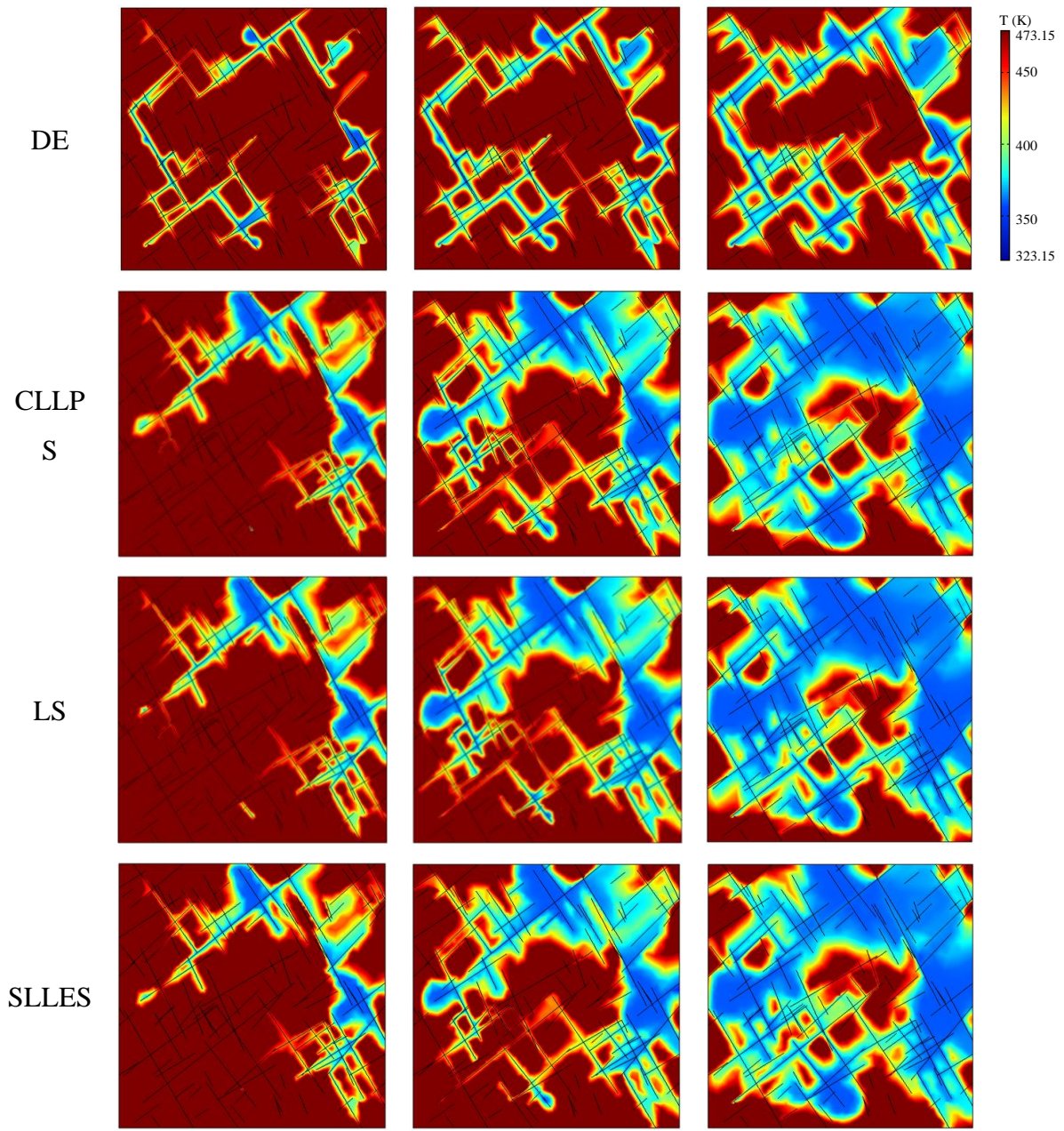

t=1500 days       t=3000 days       t=6000 days

**Fig. 11.** Thermal distributions of optimized control schemes of DE, CLLPS, LS and SLLES for the fractured geothermal reservoir after 1500, 3000 and 6000 days.

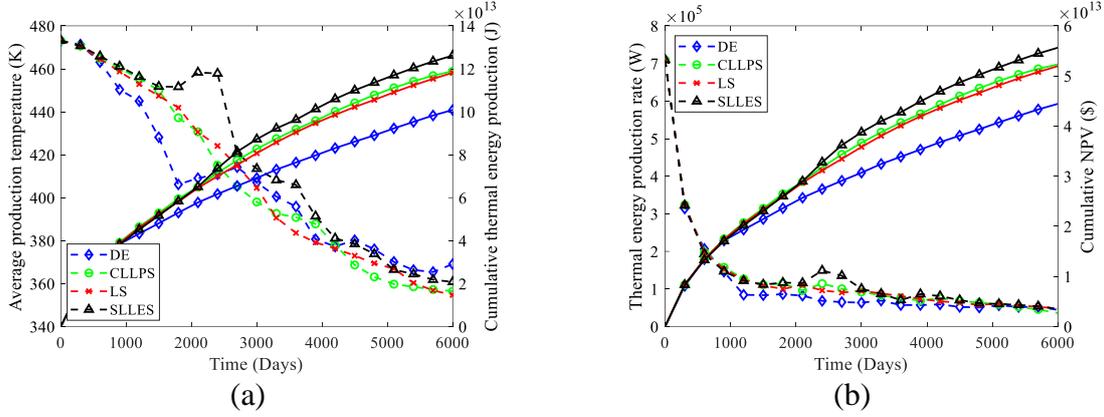

(a)                                          (b)

**Fig. 12.** The geothermal reservoir development performances obtained by DE, CLLPS, LS and SLLES for case 2.

### 5.3 Case 3: Optimization design for a field-scale EGS

To further investigate the effectiveness of SLLES on real-world geothermal exploitation problems, a field-scale EGS is designed in this study. The field-scale EGS is composed of 1264 fractures, which contains three fracture sets (**Fig. 13**). The dip direction and dip of the three fracture sets follow Fisher distribution with $\kappa = 50$. The detailed fracture distribution and parameter settings of the EGS model are presented in **Fig. 10** and **Table 6**. The size of the model is $1000 \times 500 \times 50 m$, with three injection wells and five production wells developed in the reservoir. The optimization objective is to maximize the NPV of the thermal reservoir during the project period, which equals the net energy production value minus the cost of water production and injection. The initial temperature of the field-scale EGS is 200 °C and the temperature of injection water is 50 °C. The lifetime of the geothermal reservoir project is 6000 days and each timestep is 300 days. The production wells are operated as water production rate, while the injection wells are operated as bottom hole pressure. Therefore, the goal of the EGS is to determine the control scheme of 3 injection wells and 5 production wells at 20 time steps, totally 160 variables to be determined.

**Table 6.** Parameter settings of the field-scale EGS

| Parameter | Value |
| --- | --- |
| Initial temperature | 200 °C |
| Initial pressure | 30 MPa |
| Temperature of injection water | 50 °C |
| BHP range of injection wells | [30MPa, 40 MPa] |
| Flow rate range of production wells | $[0, 20] \times 10^{-3} m^3/s$ |
| Reservoir depth | 5000m |
| Porosity of rock matrix | 0.1 |
| Porosity of fracture | 0.5 |
| Permeability of rock matrix | $5 \times 10^{-15} m^2$ |
| Permeability of fractures | $10^{-7} m^2$ |
| Reservoir thickness | 50 m |
| Thermal conductivity of rock matrix | 2 W/(m*K) |
| Thermal conductivity of water | 0.698 W/(m*°C) |
| Heat capacity of rock matrix | 850 J/(kg °C) |
| Heat capacity of water | 4200 J/(kg °C) |

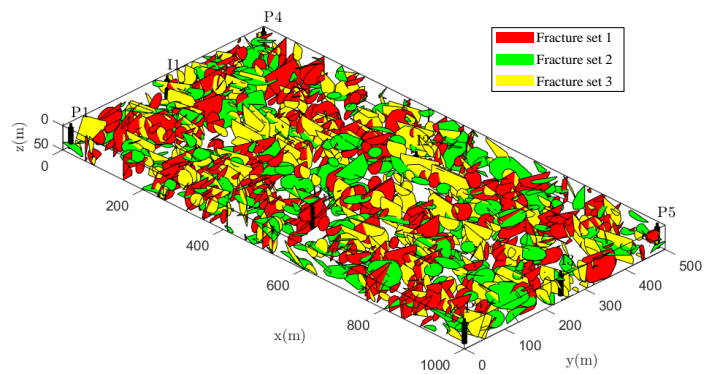

(a) DFN model with three fracture sets

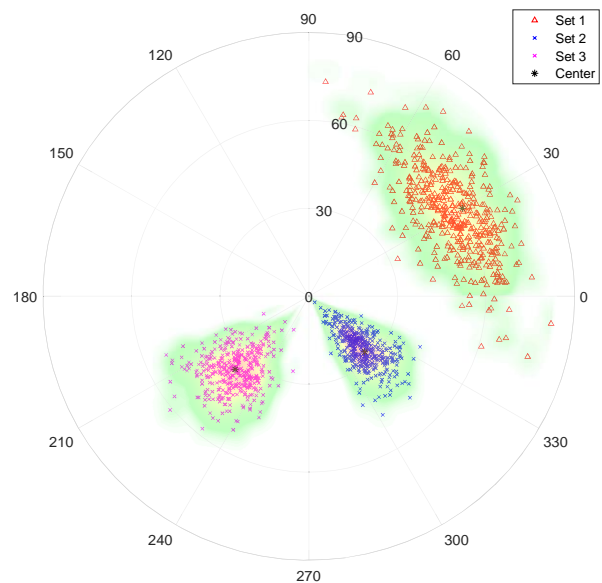

**Fig. 13.** Field-scale EGS composed of 1264 fractures: (a) DFN model with three fracture sets; (b) Polar plot of dip direction and dip distribution

For this case, the EGS model is optimized by DE, CLLPS, LS and SLLES algorithms. Since the dimension of the optimization problem is high, all surrogate-assisted algorithms sample 200 initial points into the database by LHS. Subsequently, 400 online iteratively sampling is performed. Thus, the total number of simulation evaluations is 600. Convergence curves and sampling processes of DE, CLLPS, LS and SLLES for the field-scale EGS are compared in **Fig. 14**. DE still converges slowly and CLLPS has not converged after 600 simulation evaluations. LS becomes unstable after 300 simulation evaluations because local surrogate only approximates a small local area and may lose some fidelity at other areas. LS converges efficiently at around 350 simulation evaluations although trappings into local optima, while CLLPS holds better global search ability. After taking advantage of the two sampling strategies, SLLES holds both better convergence speed and global search ability and can achieve higher NPV than DE, CLLPS and LS after 600 simulation evaluations.

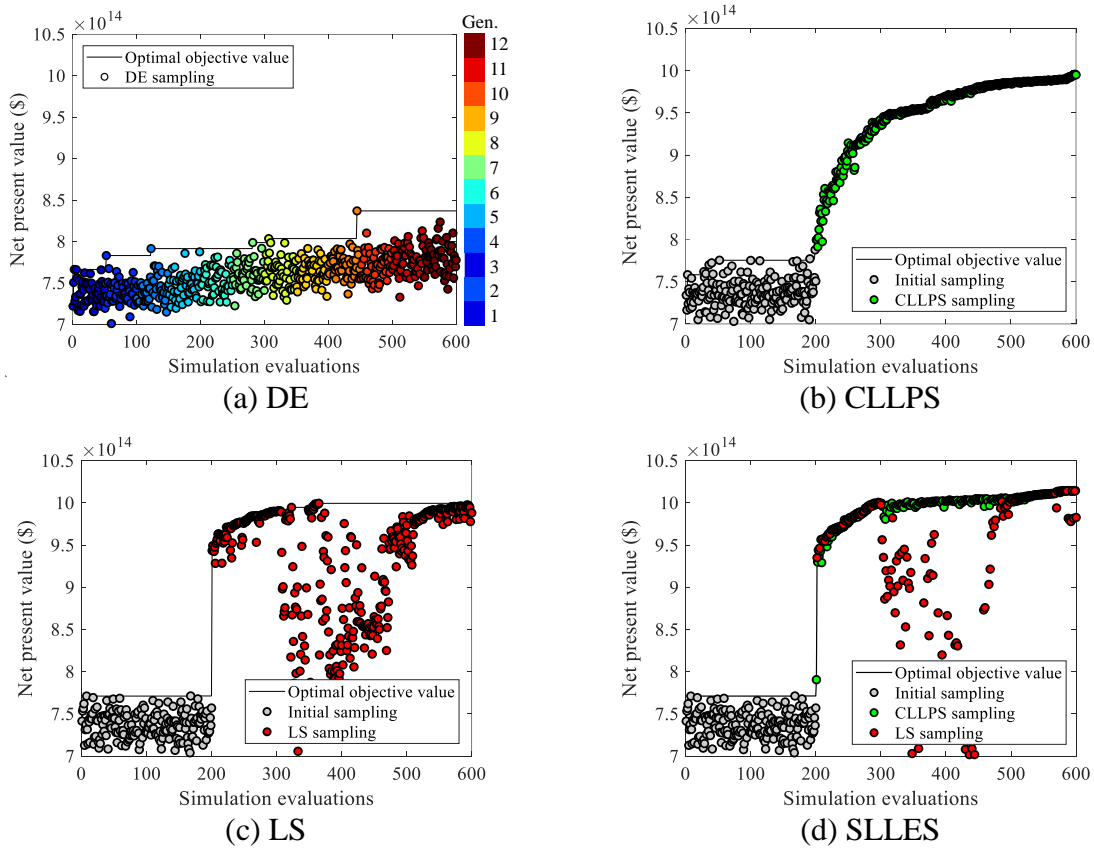

**Fig. 14.** Convergence curves and sampling processes of DE, CLLPS, LS and SLLES for the 3D field-scale EGS.

**Fig. 15** presents the optimal well-control schemes obtained by DE, CLLPS, LS and SLLES for the field-scale EGS. The bang-bang or bang-singular solution phenomenon of the well-control scheme of DE is not obvious, while the phenomenon is severe for CLLPS, LS and SLLES. Thermal distributions for optimized well-control scheme of DE, CLLPS, LS and SLLES for the fractured geothermal reservoir after 1500, 3000 and 6000 days are compared in **Fig. 16**. **Fig. 17** illustrates thermal distributions of DFN for optimized control schemes of DE, CLLPS, LS and SLLES for the fractured geothermal reservoir after 6000 days. The injected cold water mainly flows along the fracture network, displacing and carrying heat out to the production well. In comparison with DE, the thermal distributions of surrogate-assisted algorithms have larger low-temperature zone, which means that the working fluid carries more heat out to the ground.

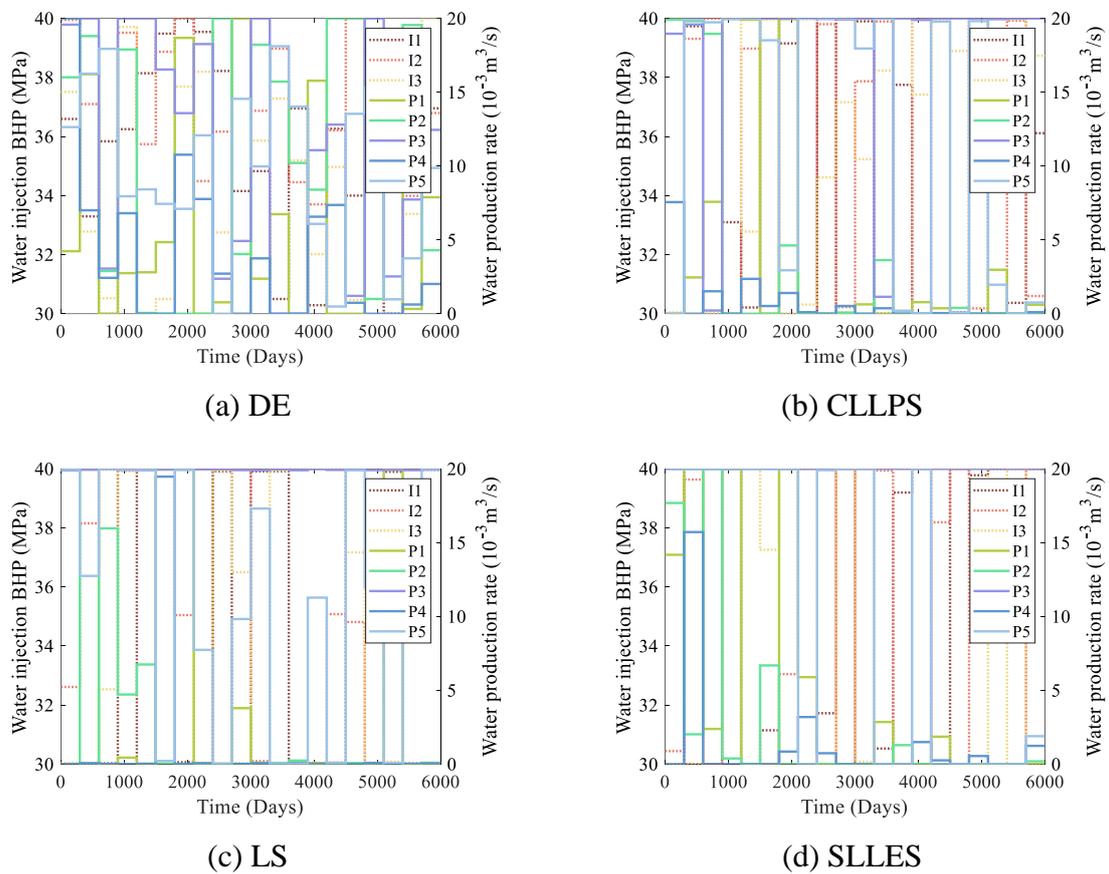

**Fig. 15.** The optimal well-control schemes obtained by DE, CLLPS, LS and SLLES for the 3D field-scale EGS.

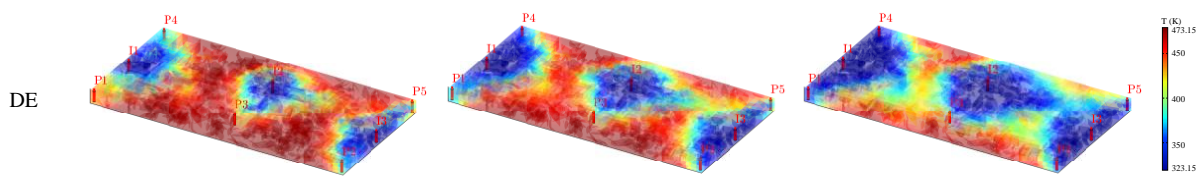

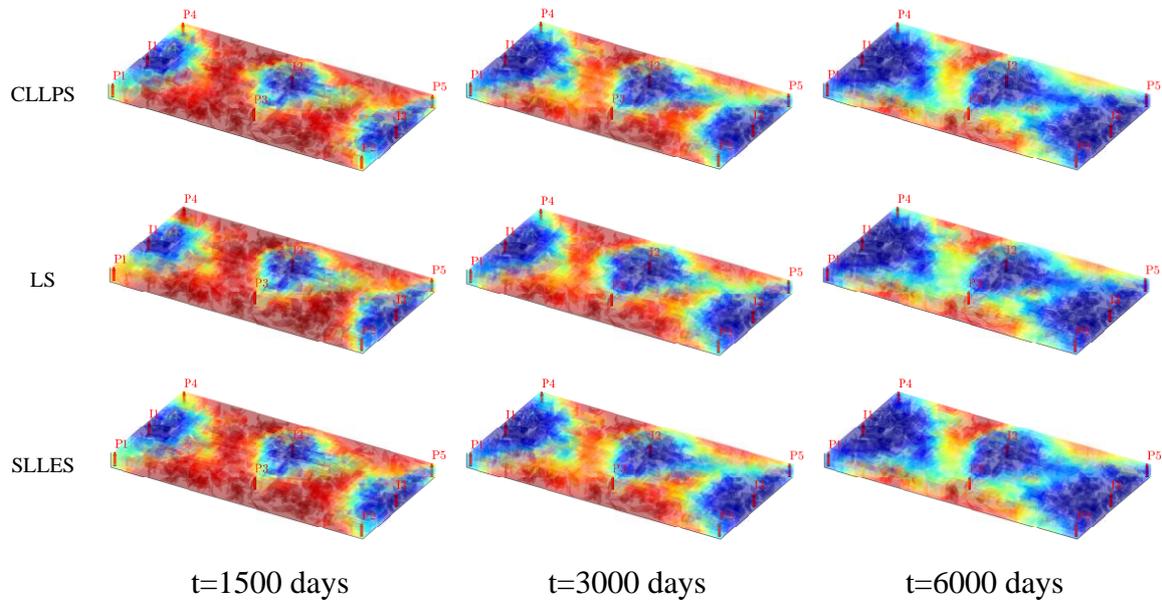

**Fig. 16.** Thermal distributions for optimized well-control schemes of DE, CLLPS, LS and SLLES for the fractured geothermal reservoir after 1500, 3000 and 6000 days.

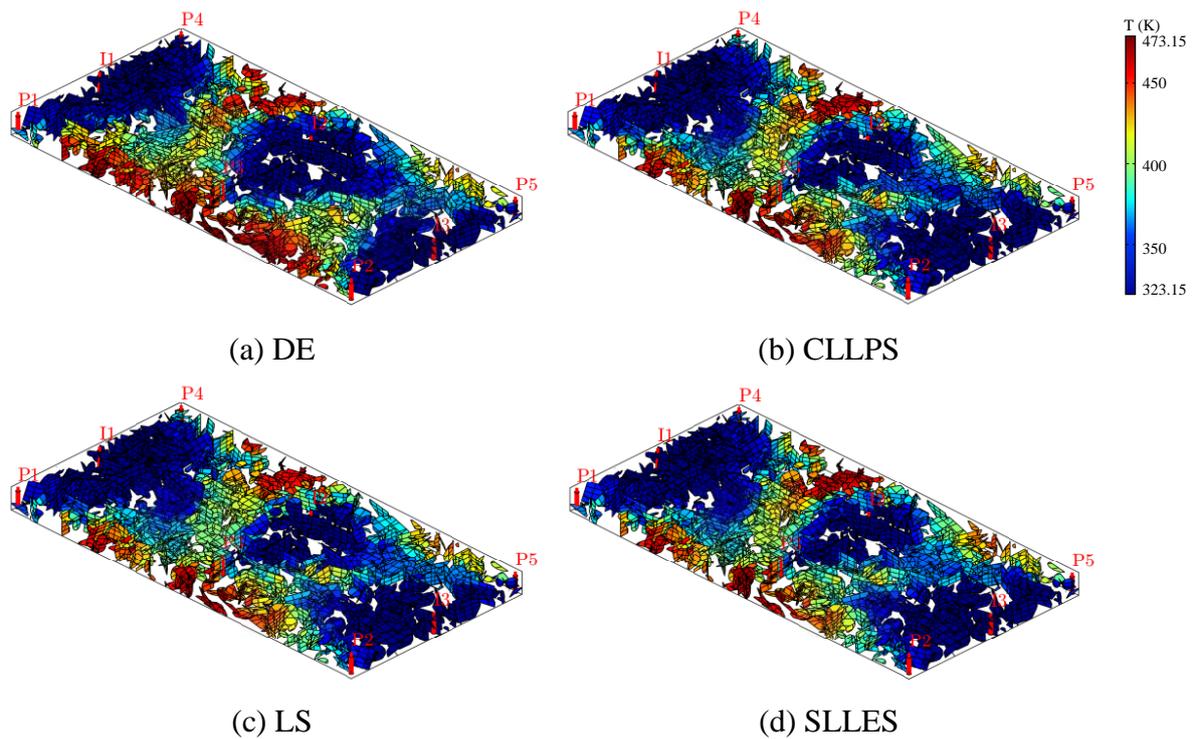

**Fig. 17.** Thermal distributions of DFN for optimized well-control schemes of DE, CLLPS, LS and SLLES for the fractured geothermal reservoir after 6000 days.

Quantitative information is hard to obtain from the temperature distribution of fracture networks and geothermal reservoirs in **Fig. 16** and **Fig. 17**. Alternatively, the geothermal reservoir development performances provided by DE, CLLPS, LS and SLLES for case 3 are evaluated and compared (**Fig. 18**). Average production temperature, cumulative thermal energy production, NPV rate and cumulative NPV of the algorithms are further illustrated. The heat extraction performance of DE is worse than other surrogate-assisted algorithms in all four criteria. The well-control scheme of SLLES achieves better performance on the cumulative thermal energy production and NPV. Although comparable in cumulative thermal energy production to SLLES, LS shows a lower cumulative NPV after 6000 days due to relatively high water-injection and -production costs. The average production temperature and NPV rate of the EGS for all four algorithms show an overall downward trend. In comparison with other methods, SLLES maintains relatively high NPV rate throughout the production period.

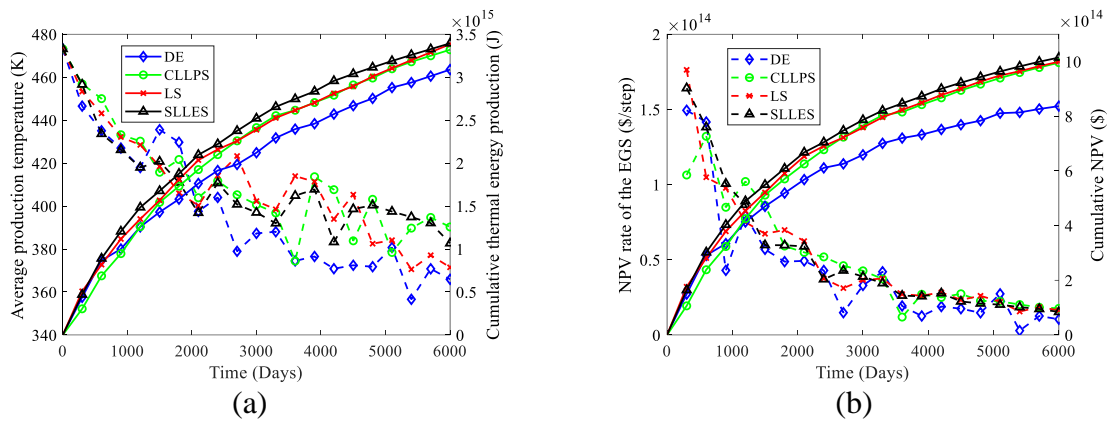

**Fig. 18.** The geothermal reservoir development performances obtained by DE, CLLPS, LS and SLLES for case 3 (Dashed line: left axis; solid line: right axis).

## 6. Discussion

As the prerequisite of heat extraction optimization design, fracture characterization and modeling for EGS remains challenging. Fusing multi-source data, which requires high density of information source from geophysical information to *in-situ* experiments and laboratory tests, pave the way to improve fracture description and modelling. However, there are many fracture distribution patterns with different geological structures and intrinsic properties, which can hardly be constrained by limited prior information. How to fully leverage prior information and observing data obtained by hydraulic tomography [23], stress-based tomography [24], tracer test [25], thermal experiments [13] and geophysical signals [26] is essential to reduce the uncertainty of fracture distribution for future thermal distribution prediction. Deep learning provides powerful performance on surrogate modelling and parameterization for the inference of complex geological structures and non-Gaussian fracture and permeability fields [37, 60, 61]. The application of deep learning in parameterizing space variations of fracture density deserves further exploration.

Geothermal energy production optimization under uncertainty of fracture distribution of EGS is also a critical issue to obtain robust control schemes and reduce the development risks [5]. Due to insufficient understanding of the subsurface fractured geothermal reservoir, the parameters of the fracture distribution of model are uncertain, which may lead to the uncertainty in geothermal prediction. Making decision under such fracture distribution uncertainties is much computationally expensive, as each evaluation of objective function needs simulations for multiple models. How to use surrogate model to reduce the number of simulation evaluations and apply transfer learning between models to learn the interaction between fracture network realizations requires further in-depth investigation.

Besides heat extraction problems, the proposed optimization workflow can also be applied to several other energy systems, especially for simulation-involved optimization design problems. For example, wind turbine airfoil design optimization is considered to maximize lift-to-drag ratio and lift coefficient of the airfoil [62]. The blade geometry optimization is developed to maximize annual energy production performance while taking structure, noise, and the cost into consideration [51]. Well placement and well-control scheme of oil reservoir need to be decided to improve the production performance and maximize the cumulative oil production [44]. The optimization of the site selection of solar power plants is aimed at achieving techno-economic optimization, thus, maximizing the reliability of the supply and minimizing costs [63]. The goal of geologic $CO_2$ sequestration optimization is to determine the optimal monitoring well locations to maximize the storage while reducing the risk of $CO_2$ leakage [55]. The proposed optimization framework is able to provide efficient and robust optimization performance in solving the above problems.

## 7. Conclusions

In this work, we proposed a novel surrogate-assisted level-based learning evolutionary search algorithm called SLLES for heat extraction optimization of EGS. SLLES takes advantages of both the level-based learning strategy and the classification model into account with the combination of evolutionary optimizer to improve the robustness and scalability of SAEAs in dealing with high-dimensional simulation-involved optimization problems. SLLES consists of classifier-assisted level-based learning pre-screen part and local evolutionary search part. Concretely, the classifier-assisted level-based learning strategy uses PNN as classifier to classify the offspring into pre-set number of levels. The offspring in different levels use level-based learning operation to generate more promising and informative candidates pre-screened by classifier to conduct hydro-geothermal simulation evaluations. For the local evolutionary search, RBF surrogate is constructed at the local promising region near the current optima. The optima of the surrogate model located by the optimizer is selected to conduct hydro-thermal simulation evaluations. The cooperation of the two strategies is able to balance the exploration and exploitation during the entire optimization process. After iteratively sampling from the design space, the robustness and effectiveness of the algorithm can be improved significantly.

Comparative experiments are systematically conducted on four benchmark functions, a 2D theoretical fractured reservoir and a 3D field-scale EGS. The results illustrate that SLLES shows superior optimization performance compared to traditional evolutionary algorithms and other recently proposed state-of-the-art surrogate-assisted algorithms. This study lays a solid basis for efficient geothermal extraction of EGS. It is worth noting that the proposed algorithm is not limited to geothermal optimization. As a general data-driven optimization framework, the developed workflow sheds light on the model management strategies of optimization in the areas of energy exploitation for complex system involving time-consuming simulation prediction. The proposed optimization framework shows potential applications in other energy systems, e.g., wind turbine performance optimization, solar photovoltaic energy optimization, hydrocarbon resource production optimization and $CO_2$ sequestration storage.

For potential future study, uncertainty in fracture distribution will be considered in decision making. The proposed algorithm can be extended to a multi-objective surrogate-assisted evolutionary algorithm to maximize the profits while minimizing the variance of multiple model outputs simultaneously. Besides, how to approximate the pressure and temperature distributions using the topology of the fracture network by relying on the power of graph-based deep learning is also a worthy research direction. After fusing the mass conservation equation into the training objective of neural network, the accuracy of the surrogate can be improved significantly.

## Acknowledgement

This study is supported by grants from the HKU Seed Fund for Basic Research, and General Fund of Natural Science Foundation of Guangdong Province (No. 2019A1515110021).